\def\year{2020}\relax
\documentclass[letterpaper]{article} % DO NOT CHANGE THIS
\usepackage{aaai20}  % DO NOT CHANGE THIS
\usepackage{times}  % DO NOT CHANGE THIS
\usepackage{amsthm}
\usepackage{amssymb}
\usepackage{subcaption}
\usepackage{amsfonts}
\usepackage{amsmath}
\usepackage{array}
\usepackage{helvet} % DO NOT CHANGE THIS
\usepackage{courier}  % DO NOT CHANGE THIS
\usepackage[hyphens]{url}  % DO NOT CHANGE THIS
\usepackage{graphicx} % DO NOT CHANGE THIS
\usepackage{graphicx}  % DO NOT CHANGE THIS
\newtheorem{definition}{Definition}
\newtheorem{theorem}{Theorem}

\newtheorem{corollary}{Corollary}
\nocopyright
\title{GraLSP: Graph Neural Networks with Local Structural Patterns}
\author{\protect{Yilun Jin,\textsuperscript{\rm 1}\thanks{Work done during undergraduate study at Peking University}} \protect{Guojie Song,\textsuperscript{\rm 2}\thanks{Guojie Song is the corresponding author.}} Chuan Shi\textsuperscript{\rm 3}\\
\textsuperscript{\rm 1}The Hong Kong University of Science and Technology, Hong Kong SAR, China\\
\textsuperscript{\rm 2}Key Laboratory of Machine Perception, Ministry of Education, Peking University, China\\
\textsuperscript{\rm 3}Beijing University of Posts and Telecommunications, China\\
yilun.jin@connect.ust.hk, gjsong@pku.edu.cn, shichuan@bupt.edu.cn
}
 
\begin{document}

\maketitle
\begin{abstract}
It is not until recently that graph neural networks (GNNs) are adopted to perform graph representation learning, among which, those based on the aggregation of features within the neighborhood of a node achieved great success. However, despite such achievements, GNNs illustrate defects in identifying some common structural patterns which, unfortunately, play significant roles in various network phenomena. In this paper, we propose GraLSP, a GNN framework which explicitly incorporates local structural patterns into the neighborhood aggregation through random anonymous walks. Specifically, we capture local graph structures via random anonymous walks, powerful and flexible tools that represent structural patterns. The walks are then fed into the feature aggregation, where we design various mechanisms to address the impact of structural features, including adaptive receptive radius, attention and amplification. In addition, we design objectives that capture similarities between structures and are optimized jointly with node proximity objectives. With the adequate leverage of structural patterns, our model is able to outperform competitive counterparts in various prediction tasks in multiple datasets.
\end{abstract}
\section{Introduction}
Graphs are ubiquitous due to their accurate depiction of relational data. Graph representation learning \cite{cui2018survey}, in order to alleviate sparsity and irregularity of graphs, came into life, projecting nodes to vector spaces while preserving graph properties. Vector spaces being regular, graph representation learning hence serves as a versatile tool by accommodating numerous prediction tasks on graphs. 

More recently, success in extending deep learning to graphs brought about Graph Neural Networks (GNNs) \cite{zhou2018graph} and achieved impressive performances.
Many GNNs follow a recursive scheme called \textit{neighborhood aggregation}, where the representation vectors of nodes are computed via aggregating and transforming the features within their neighborhoods.
By doing so, a computation tree is constructed which is computed in a bottom-up manner.
Being powerful yet efficient, neighborhood aggregation based GNNs\footnote{In this paper we focus on GNNs based on neighborhood aggregation, like \cite{xu2018powerful}, and leave other architectures for future work.} have hence attracted the attention of numerous research works  \cite{xu2018powerful,liu2019geniepath}.

In addition to node-level features which GNNs aggregate, features of other scales also prevail in graphs, among which, structural patterns of varying scales recurring frequently are typical and indicative of node and graph properties, such as functions in molecular networks \cite{prvzulj2007biological}, pattern of information flow (Granovetter \shortcite{granovetter1977strength}, Paranjape \shortcite{paranjape2017motifs}), and social phenomena \cite{kovanen2013temporal}, which are often global insights which node-level features fail to provide. 

Yet, although GNNs do encode neighborhoods of nodes \cite{xu2018powerful}, they are not ensured to generate distinctive results for nodes with different structural patterns. Specifically, distinctions between local structural patterns are minuscule, one or two links for example, which makes it hard for GNNs to generate distinctive embeddings for structural patterns, even with wildly different semantics.

\begin{figure}[h]
    \centering
    \pdfimageresolution=300 \includegraphics[width=0.95\columnwidth]{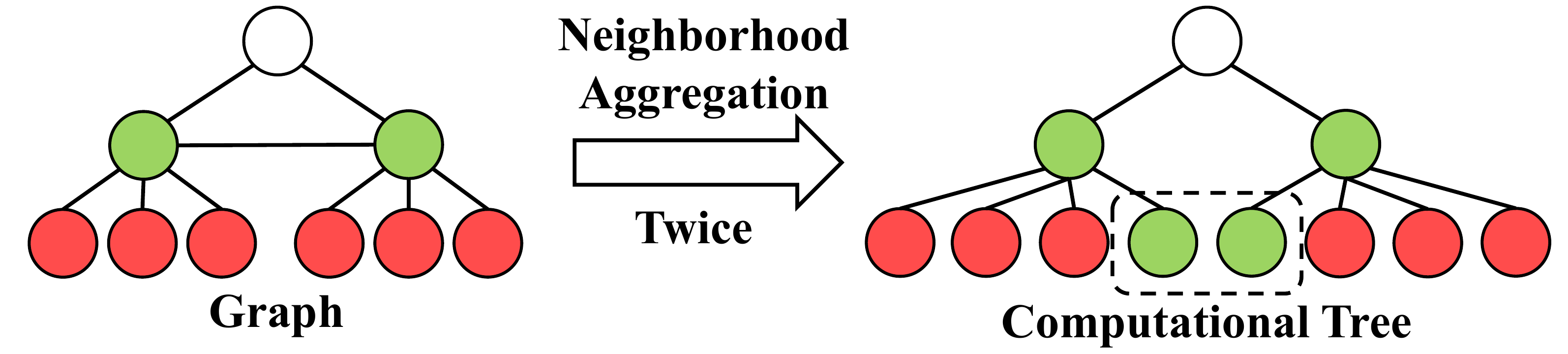}
    \caption{Computational Tree of a Triadic Closure Graph}
    \label{fig:comp_tree_triad}
\end{figure}

We take the triadic closures, patterns characteristic of strong ties in social networks, as examples \cite{huang2015triadic}. We show the computation tree of a triadic closure in a 2-layer GNN in Fig. \ref{fig:comp_tree_triad}. As can be shown, the only difference the triadic closure makes is the existence of first order neighbors (green nodes) on the second layer of the tree, whose impact tends to diminish as their neighborhood (red nodes) expands. It is thus concluded that, based on neighborhood aggregation, GNNs can, in some cases, fail to generate distinctive embeddings for structural patterns that are topologically similar but carry wildly different semantics.

As the counterpart of CNNs in images, we would consider GNNs to be capable of capturing graphical features of varying levels and scales, including both node and local, structural level. Hence, one question arises: how can we enable GNNs to more adequately capture and leverage multi-scaled structural and node features? 
One straightforward way is to first measure the structural properties of each node and concatenate them with their node features as inputs. Yet easy as it is, two challenges remain to be solved. 

\begin{itemize}
    \item \textbf{Efficiency}. Most metrics of measuring structural patterns require enumeration and pattern matching, which would often require very high complexity. For example, as one widely adopted metric, it would take an $O(|V|d^{k-1})$ time complexity to compute the $k$-graphlet statistics \cite{shervashidze2009efficient} within a graph.%, where $|V|$ is the number of nodes and $d$ is the maximum node degree. 
    \item \textbf{Incorporation of structural properties}. Challenges also lie in the incorporation of such properties. On one hand, structural features convey rich semantics that shed light on graph properties, which cannot be captured by statistics only. On the other hand, structural properties may indicate roles of nodes in graphs and hence guide the aggregation of features \cite{liu2019geniepath,ying2018graph}.
 \end{itemize}

Consequently, to complement GNNs for better addressing these structural patterns, we propose \textit{Graph Neural Network with Local Structural Patterns}, abbreviated \textbf{GraLSP}, a GNN framework incorporating local structural patterns into the aggregation of neighbors. Specifically, we capture local structural patterns via \textit{random anonymous walks}, variants of random walks that are able to capture local structures in a general manner. The walks are then projected to vectors to preserve their underlying structural semantics. We design neighborhood aggregation schemes with multiple elaborate techniques to reflect the impact of structures on feature aggregation. %The vectors for walks are then fed into the neighborhood aggregation with \textit{attention} and \textit{amplification} modules to address the interaction between structures and node contents. 
In addition, we propose objectives to jointly optimize the vectors for walks and nodes based on their pairwise proximity. Extensive experiments show that due to our elaborate incorporation of structural patterns, our model outperforms competitive counterparts in various tasks. 

To summarize, we make the following contributions. 
\begin{itemize}
    \item We analyze the neighborhood aggregation scheme and conclude that common GNNs suffer from defects in identifying some common structural patterns. 
    \item We propose that random walks can be used to capture structural patterns with analyses on them.
    \item We propose a novel neighborhood aggregation scheme that combines the structural and node properties through adaptive receptive radius, attention and amplification. 
    \item We carry out extensive experiments with their results showing that our model, incorporating structural patterns into GNNs, attains satisfactory performances.
\end{itemize}

\section{Related Work}
\textbf{Graph Representation Learning (GRL)}. Transforming discrete graphs to vectors, GRL has become popular for tasks like link prediction \cite{chen2018pme}, community detection \cite{wang2017community,long2019hierarchical} etc. 

There are generally two types of GRL methods, as defined by different notions of node similarity. On one hand, methods like DeepWalk \shortcite{perozzi2014deepwalk} and GraphSAGE \shortcite{hamilton2017inductive} adopt the notion of \textbf{homophily}, similarity defined by close connections. On the other hand, methods like struc2vec \shortcite{ribeiro2017struc2vec} and Graphwave \shortcite{donnat2018learning} define similarity as possessing similar topological structures. It should be noticed that although our method captures structural patterns, it, like most GNNs, falls into the former type, adopting the idea of homophily instead of structural similarity. We will demonstrate more on the two notions of node similarity in the experiments. 

\noindent \textbf{Graph Neural Networks (GNNs)}. GNNs (Scarselli \shortcite{scarselli2008graph}, Bruna \shortcite{bruna2013spectral}, Niepert \shortcite{niepert2016learning}, Kipf \shortcite{kipf2016semi}) gradually gain tremendous popularity in recent years.
Recent researchers generally adopt the method of neighborhood aggregation, i.e. merging node features within neighborhoods to represent central nodes (Hamilton \shortcite{hamilton2017inductive}). 

Identifying the connection between GNNs and graph structures have also been popular. \cite{xu2018powerful} and \cite{morris2019weisfeiler} demonstrated the equivalence between GNNs and the 1-WL isomorphism test. \cite{li2018deeper} showed the connection between GNN and Laplacian smoothing. Compared to previous works, our work focus more on ``local'' structures while \cite{xu2018powerful} focus more on global graph structures, e.g. graph isomorphism. 

\noindent \textbf{Measuring Structural Patterns}. Previous works on measuring structural patterns pay their attention on characteristic structures including shortest paths and graphlets (Shervashidze \shortcite{shervashidze2009efficient}, Borgwardt \shortcite{borgwardt2005shortest}) etc. %Yet almost all of them suffer from a complexity too high to be affordable in representation learning, such as $O(|V|d^{k-1})$ \cite{shervashidze2009efficient}. 
In addition, \cite{micali2016reconstructing} showed that it is possible to reconstruct a local neighborhood via anonymous random walks on graphs, a result surprising and inspiring to our model. Such notions of anonymous walks are extended by \cite{ivanov2018anonymous} who proposed graph embedding methods on them. 

\begin{figure*}[!htbp]
    \centering
    \pdfimageresolution=300 \includegraphics[width=0.95\textwidth]{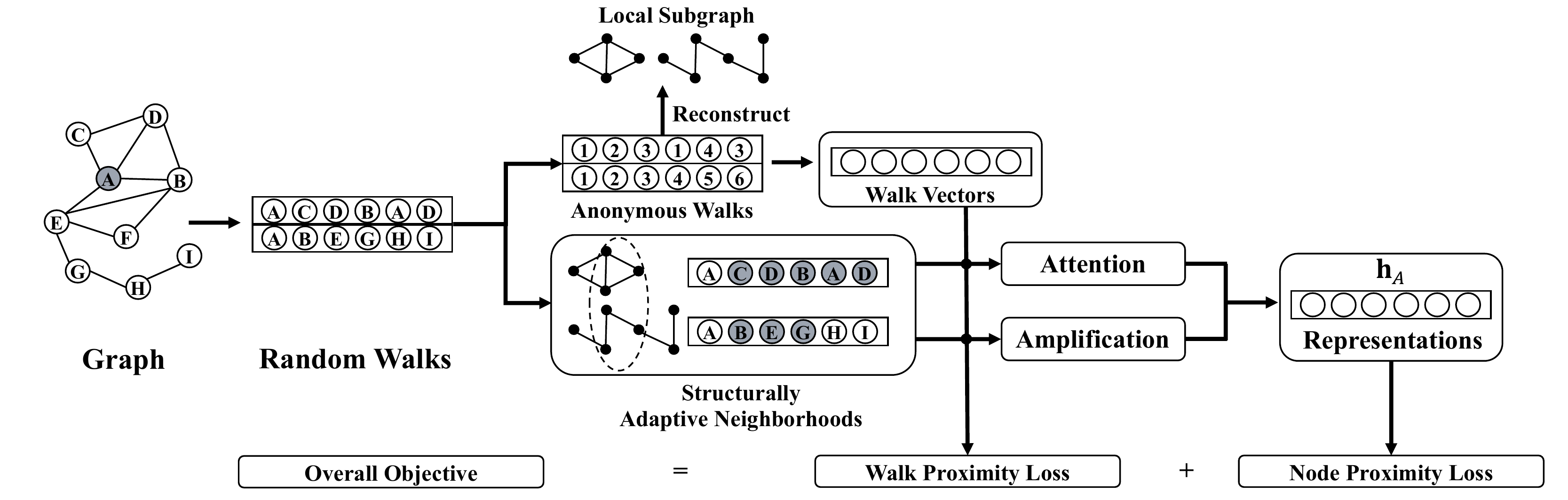}
    \caption{Overview of our model, GraLSP. For a certain node, we sample random anonymous walks around it. Anonymous walks are projected to vectors which are then aggregated along a structurally aware neighborhood via attention and amplification. The model is optimized via a joint loss of both structural and node proximity. }
    \label{fig:overview}
\end{figure*}

\section{Model: GraLSP}
In this section we introduce the design of our model, \textbf{GraLSP}, with a brief overview illustrated in Fig. \ref{fig:overview}.

\subsection{Preliminaries}
We begin by introducing several backgrounds related to our problem, including graph representation learning, random and anonymous walks, and graph neural networks. 
%\subsection{Definitions}
\begin{definition}[Graph \& Graph Representation Learning]
Given a graph $G = ( V, E)$, where $V = \{v_1, ...v_{|V|}\}$ is the set of nodes and $E = \{\langle v_i, v_j\rangle\}$ is the set of edges, graph representation learning learns a mapping function 
\begin{align*}
f: V &\to \mathbb{R}^d\\
v_i &\mapsto \mathbf{h}_i
\end{align*}
where $d\ll |V|$ with $\mathbf{h}_i$ maintaining properties of node $v_i$.
\end{definition}

Specifically, GNNs characterize their mapping functions to be iterative, where a node's representation vector is computed via aggregation of features within its neighborhood, which can be summarized by the following equation
\begin{align}
    \mathbf{h}_i^{(k)} &=\mathrm{AGGREGATE}\left(\left\{\mathbf{h}_j^{(k-1)}, v_j \in N(v_i)\cup \{v_i\}\right\}\right).
    \label{eqn:gnn}
\end{align}

Many popular GNNs, including GCN and GraphSAGE can be generalized by Eqn. \ref{eqn:gnn} \cite{xu2018powerful}. We then present the definition of random anonymous walks.

\begin{definition}[Random Anonymous Walks \cite{micali2016reconstructing}]
Given a random walk $\mathbf{w} = (w_1, w_2, ..., w_l)$ where $\langle w_i, w_{i+1}\rangle \in E$, the anonymous walk for $\mathbf{w}$ is defined as 
$$\mathrm{aw}(\mathbf{w}) = \left(\mathrm{DIS}\left(\mathbf{w}, w_1\right), \mathrm{DIS}\left(\mathbf{w}, w_2\right), ...\mathrm{DIS}\left(\mathbf{w}, w_l\right)\right)$$
where $\mathrm{DIS}(\mathbf{w}, w_i)$ denotes the number of distinct nodes in $\mathbf{w}$ when $w_i$ first appears in $\mathbf{w}$, i.e. 
$$\mathrm{DIS}(\mathbf{w}, w_i) =|\{w_1, w_2, ...w_p\}|,\ p = {\min}_j\{w_j = w_i\}.$$
\end{definition}
We denote anonymous walks of length $l$ as $\omega_1^l, \omega_2^l...$ according to their lexicographical order. For example, $\omega_1^4 = (1, 2, 1, 2), \omega_2^4 = (1, 2, 1, 3), \omega_3^4 = (1, 2, 3, 1)$, etc. 

The key difference between anonymous walks and random walks is that, anonymous walks depict the underlying ``patterns'' of random walks, regardless of the exact nodes visited. For example, both $\mathbf{w_1} = (v_1, v_2, v_3, v_4, v_2)$ and $\mathbf{w_2} = (v_2, v_1, v_3, v_4, v_1)$ correspond to the same anonymous walk $\mathrm{aw}(\mathbf{w}_1) = \mathrm{aw}(\mathbf{w}_2) = (1, 2, 3, 4, 2)$, even though $\mathbf{w}_1$ and $\mathbf{w}_2$ visited different nodes. 

\subsection{Extracting Structural Patterns}
We start by introducing our extraction of structural patterns through anonymous walks. For each node $v_i$, a set of $\gamma$ random walk sequences $\mathcal{W}^{(i)}$ of length $l$ are sampled. Alias sampling is used such that the sampling complexity would be $O(\gamma Vl)$. We then compute the empirical distribution of their underlying anonymous walks as 
\begin{equation}
\hat{p}(\omega_j^l|v_i) = \frac{\sum_{\mathbf{w}\in\mathcal{W}^{(i)}} \mathbb{I}(\mathrm{aw}(\mathbf{w}) = \omega_j^l)}{\gamma}. 
\end{equation}
%which, as \cite{micali2016reconstructing} has shown, is indicative of local neighborhood structure of $v_i$.
In addition, we take the mean empirical distribution over the whole graph $G$ as
\begin{equation}
\hat{p}(\omega_j^l|G) = \frac{\sum_{i=1}^{|V|}\hat{p}(\omega_j^l|v_i)}{|V|}, 
\end{equation}
as estimates of the true distribution $p(\omega_j^l|v_i)$ and $p(\omega_j^l|G)$ \cite{shervashidze2009efficient}. 

\subsubsection{Rationale of Anonymous Walks}
There are works exploring properties of anonymous walks. Micali \shortcite{micali2016reconstructing} showed that one can reconstruct a local sub-graph using anonymous walks. We present the theorem here. 
\begin{theorem}
\cite{micali2016reconstructing} Let $B(v, r)$ be the subgraph induced by all nodes $u$ such that $dist(v, u)\le r$ and $\mathcal{D}_l$ be the distribution of anonymous walks of length $l$ starting from $v$, one can reconstruct $B(v, r)$ using $(\mathcal{D}_1, ...\mathcal{D}_l)$ where $l = 2(m+1)$ and $m$ is the number of edges in $B(v, r)$.
\end{theorem}
This theorem underscores the ability of anonymous walks to capture structures in a highly general manner, in that they capture the complete $r$-hop neighborhood \footnote{Although we do not explicitly reconstruct $B(v, r)$, such theorem demonstrates the ability of anonymous walks to represent structural properties.}. Yet this theorem is unrealistic considering representing structural patterns in GNNs. For example, for the dataset Cora and $r = 2$, we get $\bar{l} = 118$, which is impossible to deal with since the number of anonymous walks grows exponentially with $l$ \cite{ivanov2018anonymous}. Instead, we propose an alternative that is more suitable for our task. 
\begin{corollary}
One can reconstruct $B^c(v, r)$ with anonymous walk of length $l = O(m + r)$ where $m$ is the number of edges in an ego-network of $G$, if one can de-anonymize the first $r-1$ elements in each anonymous walk starting from $v$. 
\end{corollary}
\begin{corollary}
Given that the graph follows power-law degree distribution, the expected number of edges $\mathbb{E}[m]$ in an ego-network of $G$ would be 
\begin{equation}
    \left(1-\frac{c}{2}\right)d + \frac{cd}{2(d-2)}\left((d_{max})^{\frac{d-2}{d-1}}-1\right)
\end{equation}
where $d, d_{max}, c$ denote average degree, maximum degree, clustering coefficient of $G$ respectively. 
\end{corollary}
These corollaries show the rationale of using reasonably long anonymous walks to depict general local structural patterns. Specifically, for citation graphs including Cora, Citeseer and AMiner, Eqn. 2 evaluates to about $10$. We omit the detailed proofs due to space constraints. 

\begin{figure}[ht]
\centering
\includegraphics[width = 0.95\columnwidth]{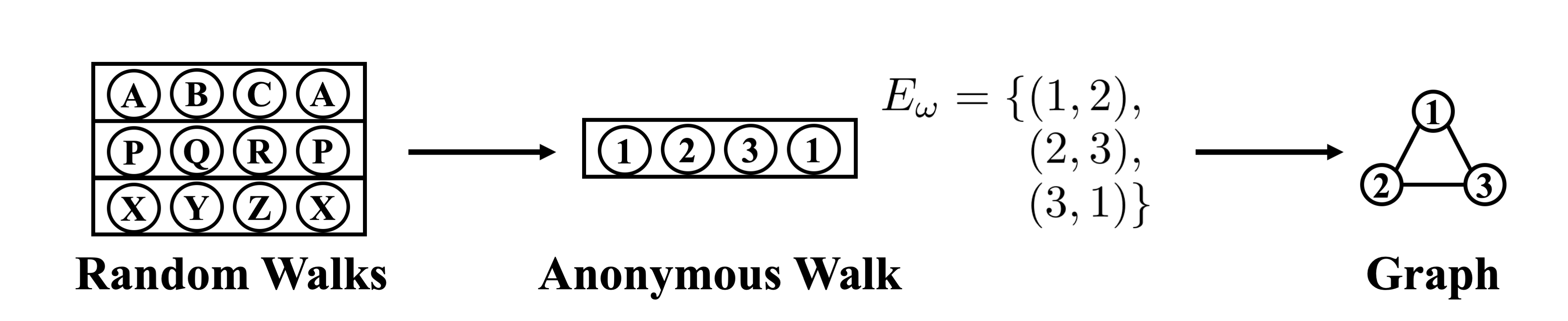}
\caption{Example of anonymous walks on a triadic closure}
\label{fig:walk_triad}
\end{figure}

In addition, we provide intuitive explanations of anonymous walks which we find appealing. Intuitively, an anonymous walk $\omega$ with $k$ distinct nodes induces a graph $G_{\omega} =( V_\omega, E_\omega)$ with $V_\omega = \{1, 2...k\}$ and $\langle i, j \rangle\in E_\omega \Leftrightarrow (j, i)$ or $(i, j)\subseteq \omega$. In this sense, a single anonymous walk is a partial reconstruction of the underlying graph, which is able to indicate certain structures, such as triads. We show the intuition with walks on a triadic closure as an example in Fig. \ref{fig:walk_triad}.

\subsection{Aggregation of Structural Patterns}
In this section we introduce our incorporation of structural patterns into the representation of nodes. 
\subsubsection{Representing Anonymous Walks}
Denoting anonymous walks as statistics is insufficient as walks represent structural patterns with varying similarities to each other. For example, we would intuitively believe that $(1, 2, 3, 1, 4)$ is highly similar to $(1, 2, 3, 1, 2)$ as they both indicate an underlying triad, but is dissimilar to $(1, 2, 3, 4, 5)$ as no triads are indicated. 

Consequently, as we would like to capture the properties of varying walk sequences, we represent each anonymous walk as a vector through an embedding table lookup
\begin{equation}
\begin{aligned}
    f^{aw}: %\{\omega_j^l\}&\to \mathbb{R}^{d'}\
    \omega_j^l &\mapsto \mathbf{u}_j^{aw}\in \mathbb{R}^{d'},
\end{aligned}
\end{equation}
to capture the properties of varying walks and structures. 

\subsubsection{Neighborhood Aggregation}
In this part we introduce how we aggregate structures along with node-level features. Specifically, we focus on how to aggregate node features under the impact of their local structural patterns, instead of plainly aggregating them together using concatenation.

Intuitively, we consider structures to have the following impacts on the aggregation of information on graphs: 
\begin{itemize}
\item \textbf{Defining Receptive Paths}. Random walks can be seen as receptive paths, showing how information flows over the graph \cite{liu2019geniepath}. Hence, we would like to define flexible receptive paths, or ``neighbors" of $v$ according to its random walks, instead of fixed 1-hop neighbors. 
\item \textbf{Importance of Neighbors}. Generally neighbors do not exert impact uniformly but exhibit varying strength. It has been studied that structures including cliques or dense clusters generally indicate strong social impact \cite{granovetter1977strength}, which should be captured by our model. 
\item \textbf{Selective Gathering of Information}. Structural patterns may also characterize selection towards the information to gather. For example, enzymes in biological networks share distinctive structures such that selective catalysis towards biological reactions is enabled \cite{ogata2000heuristic}. 
\end{itemize}

To address the above impacts, we design our aggregation formula as follows. 
\begin{align}
    \mathbf{a}^{(k)}_i &= \mathrm{MEAN}_{\mathbf{w}\in\mathcal{W}^{(i)}, p\in [1, r_\mathbf{w}]} \left(\lambda^{(k)}_{i, \mathbf{w}}\left(\mathbf{q}^{(k)}_{i, \mathbf{w}}\odot \mathbf{h}_{\mathbf{w}_p}^{(k-1)}\right)\right)\\
    \mathbf{h}^{(k)}_i &= \mathrm{ReLU}\left(\mathbf{U}^{(k)}\mathbf{h}^{(k-1)}_i + \mathbf{V}^{(k)}\mathbf{a}^{(k)}_i\right), k = 1, 2, ...K, \\
    \mathbf{h}_i &= \mathbf{h}_i^{(K)}, 
    \label{eqn:aggregation}
\end{align}
where $\mathbf{w}$ denotes a walk starting from $v_i$, $\mathbf{w}_p$ denotes the $p$-th node of walk $\mathbf{w}$. $\mathrm{ReLU}(x) = \max(0, x)$ is the ReLU activation and $\mathrm{MEAN}(\Omega) = \frac{\sum_{\omega\in \Omega}\omega}{|\Omega|}$ is the mean pooling. In addition, $\odot$ denotes element-wise multiplication, while $r_\mathbf{w}$, $\lambda_{i, w_1}, \mathbf{q}_{i, w_1}$ denote receptive radius of $\mathbf{w}$, attention and amplification coefficients, respectively, which we will introduce in detail later that correspond to the above impacts. Moreover, $\mathbf{U}^{(k)}, \mathbf{V}^{(k)}$ denote trainable weight matrices. 

\subsubsection{Adaptive Receptive Radius} While each random walk can be seen as a receptive path, properties of the walk imply different radius of reception. For example, if a walk visits many distinct nodes, it may span to nodes far away which may not exert impact on the central node. On the other hand, a walk visiting few distinct nodes indicates an underlying cluster of nodes, which are all close to the central node. Hence, we propose the adaptive receptive radius for neighborhood sampling to address it. Specifically, the receptive radius of a walk $r_\mathbf{w}$, or ``window size" is negatively correlated to its span, i.e. 
\begin{equation}r_\mathbf{w} = \left\lfloor \frac{2l}{\max(\mathrm{aw}(\mathbf{w}))}\right\rfloor,\end{equation}
where $\max(\mathrm{aw}(\mathbf{w}))$ denotes the number of distinct nodes visited by walk $\mathbf{w}$. We build the neighborhood of $v_i$ such that for each $\mathbf{w}\in \mathcal{W}^{(i)}$, only nodes within the radius $r_\mathbf{w}$ are included, which forms an adaptive neighborhood of node $v_i$.

\subsubsection{Attention}
We introduce the attention module \cite{velivckovic2017graph} to model varying importance of neighbors shown by their structural patterns. Specifically, we model $\lambda_{i, w_1}$ as follows where the notations are defined as below Eqn. \ref{eqn:aggregation}:
\begin{equation}
\lambda^{(k)}_{i, \mathbf{w}} = \frac{\exp\left(\mathbf{P^{(k)}u}^{aw}_{\mathrm{aw}(\mathbf{w})} + b^{(k)}\right)}{\sum_{\mathbf{w'}\in \mathcal{W}^{(i)}}\exp\left(\mathbf{P}^{(k)}\mathbf{u}^{aw}_{\mathrm{aw}(\mathbf{w}')} + b^{(k)}\right)},
\end{equation}
where $\mathbf{P}^{(k)}$ and $b^{(k)}$ are trainable parameters. 

\subsubsection{Amplification}
We introduce \textit{amplification} module for channel-wise amplification, or ``gate", to model the selective aggregation of node features in the neighborhood. 
Formally, we model $\mathbf{q}_{i, w_1}$ similarly as:
\begin{equation}
\mathbf{q}^{(k)}_{i, \mathbf{w}} = \sigma\left(\mathbf{Q}^{(k)}\mathbf{u}^{aw}_{\mathrm{aw}(\mathbf{w})} + \mathbf{r}^{(k)}\right),
\end{equation}
where $\sigma(x) = 1/(1+\exp(-x))$ is the sigmoid function to control the scale of amplification, and $\mathbf{Q}^{(k)}$, $\mathbf{r}^{(k)}$ are trainable parameters.

\subsection{Model Learning}
In this section we introduce the objectives guiding the learning of our model. Specifically, we design a multi-task objective function to simultaneously preserve proximity between both pairwise nodes but also pairwise walks. 

\subsubsection{Preserving Proximity of Walks} 
Intuitively, if two anonymous walks both appear frequently within the same neighborhood, they are supposed to depict similar structural information --- the same neighborhood, and vise versa. Hence, we design our walk proximity objective as follows, 
\begin{equation}
    \min L_{walk} = -\sum_{v_i\in V}\log \sigma\left((\mathbf{u}_j^{aw})^T\mathbf{u}_k^{aw} - (\mathbf{u}_j^{aw})^T\mathbf{u}_n^{aw}\right),\\
\end{equation}
\begin{align*}
    \text{s.t.}\quad \hat{p}(\omega_j^l|v_i)&>\hat{p}(\omega_j^l|G), \hat{p}(\omega_k^l|v_i)>\hat{p}(\omega_k^l|G)\\
    \hat{p}(\omega_n^l|v_i)&<\hat{p}(\omega_n^l|G)
\end{align*}
such that highly co-appearing walks are mapped with similar vectors. By constraining walk vectors in this way, we are endowing walk vectors with semantics which can interpret similarities, such that our operations of incorporating walk vectors into aggregation are sound. 

\subsubsection{Preserving Proximity of Nodes}
An objective preserving node proximity is required so as to preserve node properties. We adopt the unsupervised objective of \cite{perozzi2014deepwalk} but it does not rule out other objectives. 

\begin{equation}
    \begin{aligned}
        \min L_{node} &= -\sum_{v_i\in V}\sum_{v_j \in N(v_i)} \Big[\log\sigma\left(\mathbf{h}_i^T\mathbf{h}_j\right)\\ 
                &- k\mathbb{E}_{v_n\sim P_n(v)}\left[\log\sigma\left(\mathbf{h}_i^T\mathbf{h}_n\right)\right]\Big].
    \end{aligned}
    \label{eqn:node}
\end{equation}
%It should be noted that other objective functions can also be utilized, such as the semi-supervised cross entropy objective in GCN \cite{kipf2016semi}, 
%$
%L_{GCN} = \sum_{i=1}^{|V|}H(\mathbf{y}_i, \mathbf{z}_i).
%H(\alpha, \beta) &= \sum_j \alpha_j \log\left(\frac{1}{\beta_j}\right).
%$

\subsubsection{Overall Objective}
We combine the above two objectives together by summing them up 
\begin{equation}
    \min L_{overall} = L_{node} + \mu L_{walk}
\end{equation}
to obtain a multi-task objective preserving both proximity between pairwise walks and nodes. We adopt the Adam Optimizer to optimize the objective using TensorFlow. 

\section{Experiments}
In this section we introduce our experimental evaluations on our model GraLSP.

\subsection{Experimental Setup}

We use the following datasets for the experiments. We take nodes as papers, edges as citations, labels as research fields and word vectors as features, if not specified elsewhere. 
\begin{itemize}
    \item \textbf{Cora} and \textbf{Citeseer} are citation datasets used in GCN (Kipf \shortcite{kipf2016semi}). We reduce the feature dimensions from about 2000 to 300 and 500 through PCA, respectively. 
    \item \textbf{AMiner} is used in DANE \cite{ijcai2019-0606}. We reduce the feature dimensions from over 10000 to 1000.   
    \item \textbf{US-Airport} is the dataset used in struc2vec \shortcite{ribeiro2017struc2vec}, where nodes denote airports and labels correspond to activity levels. We use one-hot encodings as features. 
\end{itemize}
We summarize the statistics of the datasets in Table \ref{tab:dataset}.
\begin{table}[ht]
    \centering
    \resizebox{0.95\columnwidth}{!}{
    \begin{tabular}{|c|c|c |c|c|}
         \hline
         Dataset & $|V|$ & $|E|$ & Feature Dims & \# Labels\\
         \hline
         Cora & 2708 &  5429 & 300 & 7\\
         Citeseer & 3264 & 4591 & 500 & 6\\
         AMiner& 3121 & 7219 & 1000 & 4\\
         US-Airport & 1190 & 13599 & 1190 & 4\\
         \hline
    \end{tabular}}
    \caption{Dataset Statistics}
    \label{tab:dataset}
\end{table}

\begin{table*}[t]
\centering
\resizebox{1.92\columnwidth}{!}{
\begin{tabular}[width = \textwidth]{c|p{1.4cm}<{\centering}p{1.6cm}<{\centering}|p{1.4cm}<{\centering}p{1.6cm}<{\centering}|p{1.4cm}<{\centering}p{1.6cm}<{\centering}|p{1.4cm}<{\centering}p{1.6cm}<{\centering}}  
\hline
& \multicolumn{2}{c|}{Cora} & \multicolumn{2}{c|}{Citeseer} & \multicolumn{2}{c|}{AMiner} & \multicolumn{2}{c}{US-Airport}\\
&  Macro-f1 & Micro-f1 &  Macro-f1 & Micro-f1 &  Macro-f1 & Micro-f1 & Macro-f1 & Micro-f1\\
\hline
GraLSP  & \textbf{0.8412}	& \textbf{0.8518}	& \textbf{0.6458}  & \textbf{0.7041} & \textbf{0.7092}	&\textbf{0.6987} & 0.5882 & 0.5981\\
\hline
DeepWalk &  0.7053  &   0.7305    &	0.3860   & 0.4540 &   0.6157	& 0.6128 & 0.5645 & 0.5723\\
LINE & 0.5777 &0.6044	&0.3529  &0.4021 & 0.5254	& 0.5628 & 0.5560 & 0.5672\\
Struc2Vec & 0.2181 & 0.3675 & 0.2289 & 0.2907 &  0.2771 & 0.4682 & \textbf{0.6142} & \textbf{0.6285}\\
Graphwave & 0.0677 & 0.3105 & 0.1151 & 0.2469 & 0.1218 & 0.3213 & 0.5872 & 0.6128\\
GraphSAGE & 0.8169	&0.8294	&0.6213 & 0.6837	& 0.6711	&0.6603&0.5824 & 0.5941\\
GCN & 0.8131 & 0.8236 & 0.6338 & \textbf{0.7032}	&0.6567	&0.6391 & 0.0925 & 0.2269\\
GAT & 0.8166 & 0.8304 & 0.6377 & 0.6993 & 0.6392 & 0.6321 & 0.4982 & 0.5092 \\
\hline
\end{tabular}}
\caption{Macro-f1 and Micro-f1 scores of node classification on different datasets.}
\label{tab:node_classification}
\end{table*}

\begin{table*}[t]
\centering
\resizebox{1.92\columnwidth}{!}{
\begin{tabular}[width = \textwidth]{c|p{1.4cm}<{\centering}p{1.6cm}<{\centering}|p{1.4cm}<{\centering}p{1.6cm}<{\centering}|p{1.4cm}<{\centering}p{1.6cm}<{\centering}|p{1.4cm}<{\centering}p{1.6cm}<{\centering}}  
\hline
& \multicolumn{2}{c|}{Cora} & \multicolumn{2}{c|}{Citeseer} & \multicolumn{2}{c|}{AMiner} & \multicolumn{2}{c}{US-Airport}\\
&  AUC & Rec@0.5 &  AUC & Rec@0.5 &  AUC & Rec@0.5 & AUC & Rec@0.5\\
\hline
GraLSP  & \textbf{0.9465}	& \textbf{0.8834}	& \textbf{0.9577}  & \textbf{0.8957} &   \textbf{0.9659}	&\textbf{0.9131} & 0.8103 & 0.7473\\
\hline
DeepWalk &  0.8666  &   0.8055    &	0.8677   & 0.8022 & 0.9164	& 0.8525 &0.7592 & 0.7232 \\
LINE & 0.8141 &0.7664	&0.8153  &0.7512 & 0.8688	& 0.8170 & \textbf{0.9099} & \textbf{0.8209}\\
Struc2Vec & 0.6323 & 0.6022 & 0.7664 & 0.6497 & 0.6824 & 0.6269 & 0.7221 & 0.6571\\
Graphwave & 0.2999 & 0.3652 & 0.3661 & 0.4123 & 0.2875 & 0.3577 & 0.5608 & 0.5434\\
GraphSAGE & 0.9269	&0.8542	&0.9421 & 0.8776	 & 0.9484	&0.8892 & 0.8105 & 0.7435\\
GCN & 0.8779	&0.8022	& 0.8831 & 0.8048	&  0.8612	&0.7725 & 0.7126 & 0.6662\\
GAT & 0.8894 & 0.8031 &  0.8853 & 0.7956 & 0.8571 & 0.7559 & 0.7486 & 0.6933\\
\hline
\end{tabular}}
\caption{Results of link prediction on different datasets. Rec@0.5 denotes recall at 50\%. }
\label{tab:link_prediction}
\end{table*}

We take the following novel approaches in representation learning as baselines.
\begin{itemize}
    \item \textbf{Skip-gram models}, including DeepWalk \shortcite{perozzi2014deepwalk} and LINE \shortcite{tang2015line}, which optimizes proximity between nodes.
    \item \textbf{Structure models}, focusing on topological similarity instead of connections, including struc2vec and Graphwave. 
    \item \textbf{GNNs}, including GraphSAGE, GCN and GAT. We use unsupervised GraphSAGE with mean aggregator and semi-supervised GCN, GAT with 6\% labeled nodes. 
\end{itemize}

As for parameter settings, we take 32-dimensional embeddings for all methods, and adopt Adam optimizer with learning rate 0.005. For GNNs, we take 2-layer networks with a hidden layer sized 100. For models involving skip-gram optimization including DeepWalk, GraphSAGE and GraLSP, we take $\gamma = 100$, $l = 8$, window size as 5 and the number of negative sampling as 8. For models involving neighborhood sampling, we take the number for sampling as 20. In addition, we take $\mu = 0.1$, and $d' = 30$ for GraLSP, and keep the other parameters for the baselines as default. 

\subsection{Visualization as a Proof-of-Concept}
\begin{figure}[ht]
\centering
\subcaptionbox{Illustration of $\mathcal{G}(n)$}{\pdfimageresolution=300\includegraphics[width=0.7\columnwidth]{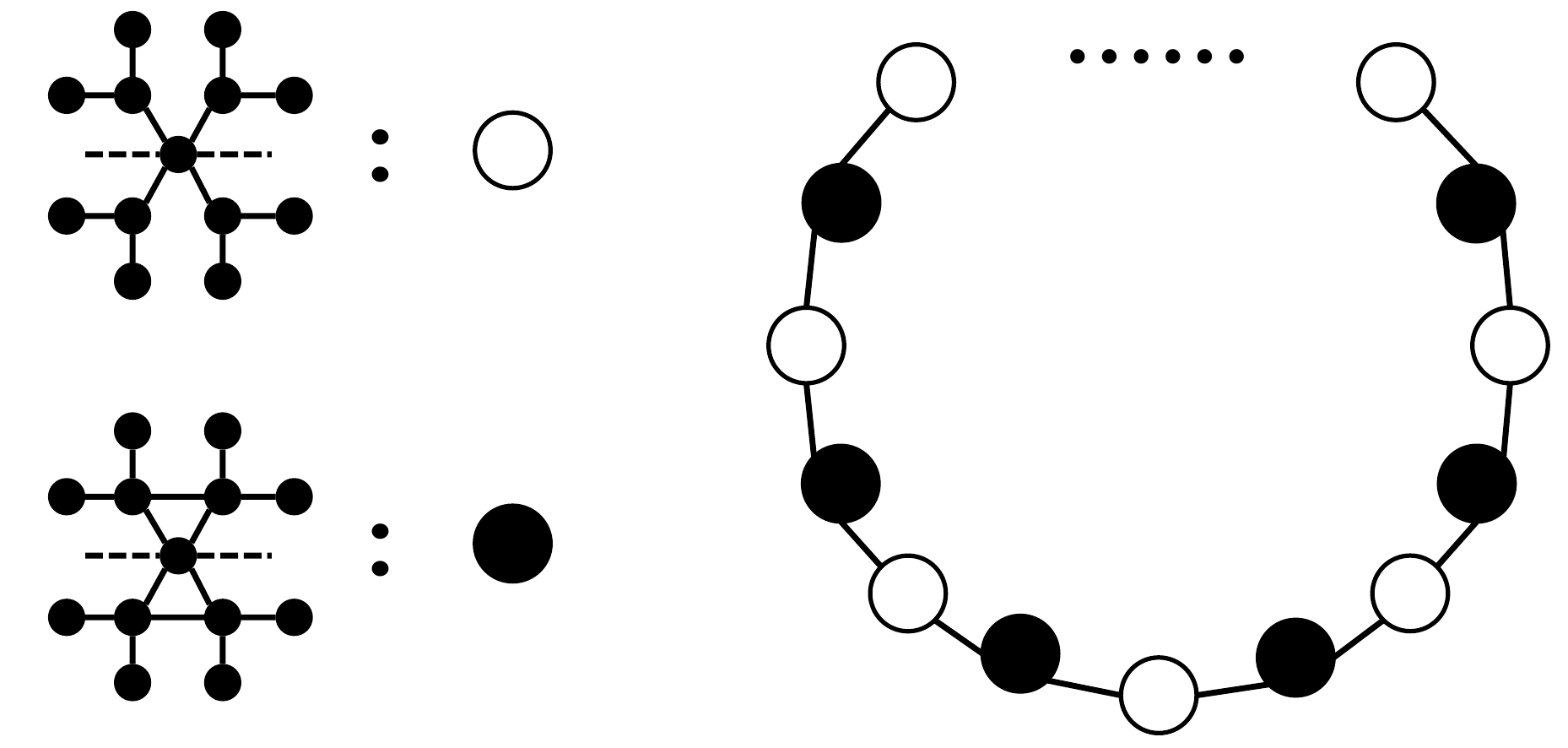}}%

\subcaptionbox{GraphSAGE}{\includegraphics[width=0.35\columnwidth]{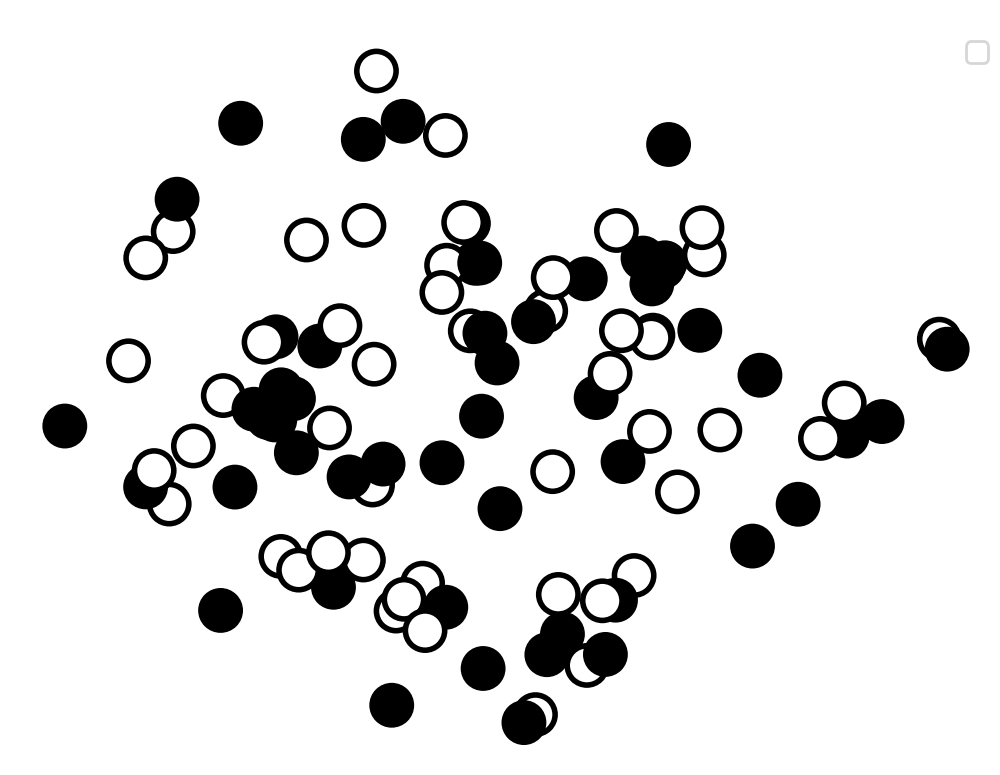}}%
\hspace{1.2cm}
\subcaptionbox{GraLSP}{\includegraphics[width=0.35\columnwidth]{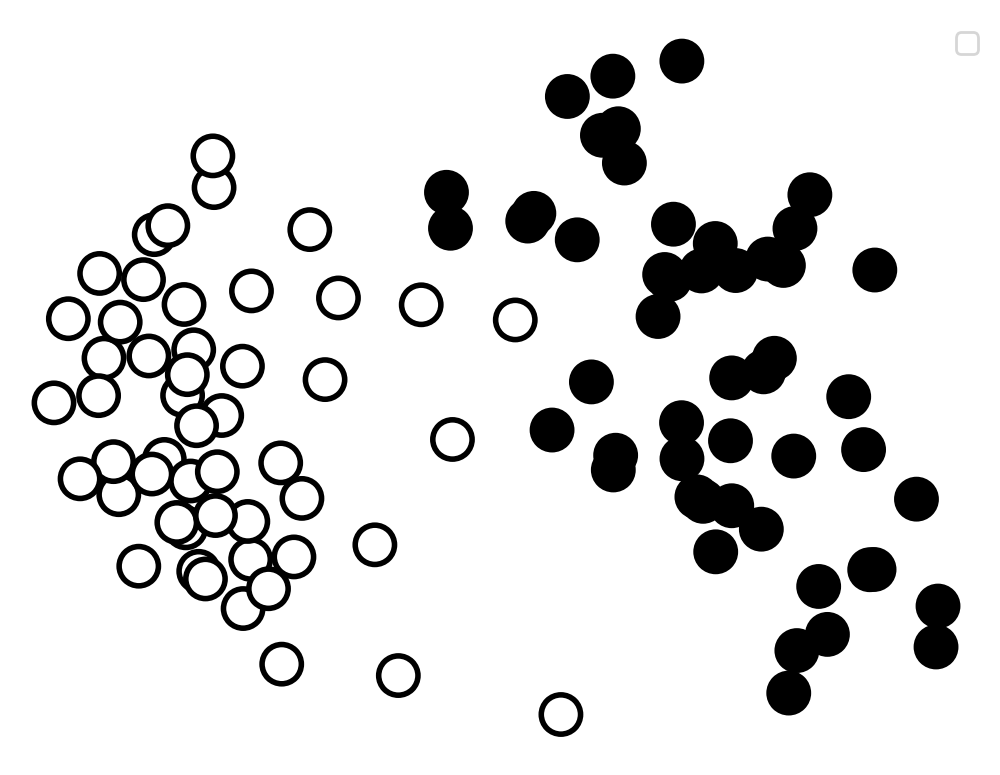}}%
\caption{Visualization on artificial graph $\mathcal{G}(100)$ where  black dots denote nodes surrounded by closed triads and white ones denote those by open triads.}
\label{fig:artificial}
\end{figure}

We first carry out visualization on an artificial dataset $\mathcal{G}(n)$ as a proof-of-concept, to test GNNs' ability to identify local structural patterns. We build $\mathcal{G}(n)$ from a circle with $n$ nodes, where each node is surrounded by either two open or closed triads interleavingly. In addition, for each triad, there are 4 addition nodes linked to it. Apparently the nodes on the circle possess two distinct structural properties, those surrounded by closed triads and those by open ones. We show the illustration of $\mathcal{G}(n)$ and its building blocks in Fig. \ref{fig:artificial}.

We visualize the representations from GraphSAGE and GraLSP in Fig. \ref{fig:artificial}. As shown, GraLSP generates a clearer boundary between the two types of nodes, while GraphSAGE fails to draw a boundary as distinctive, which not only demonstrates the inability of current GNNs in generating distinctive embeddings for different local structural patterns, but also underscores the ability of anonymous walks and GraLSP in complementing such drawbacks.

\subsection{Node Classification}
We carry out node classification on the four datasets. We learn representation vectors using the whole graph, which are then fed into \textit{Logistic Regression} in Sklearn. We take 20\% of all nodes as the test set and 80\% as training. We take the macro and micro F1-scores for evaluation. In addition, all results are averaged for 10 independent experiments. 

The results are shown in Table \ref{tab:node_classification}. As shown, the performance gain from original GNNs towards GraLSP is considerable, which demonstrates GraLSP is able to complement the drawbacks of identifying local structures. In addition, struc2vec and Graphwave perform poorly on academic datasets, but impressively on US-Airport, which can be attributed to the label definitions. In academic datasets, labels are defined as fields, where connected papers tend to have the same field and label, while in US-Airport, labels are taken as activity levels with less significant homophily but more related to structural properties. Nonetheless, we can see that generally GraLSP produces satisfactory results. 

\begin{figure*}[t]
\centering
\subcaptionbox{Walks per node \label{fig:param_gamma}}{\includegraphics[width=0.18\textwidth]{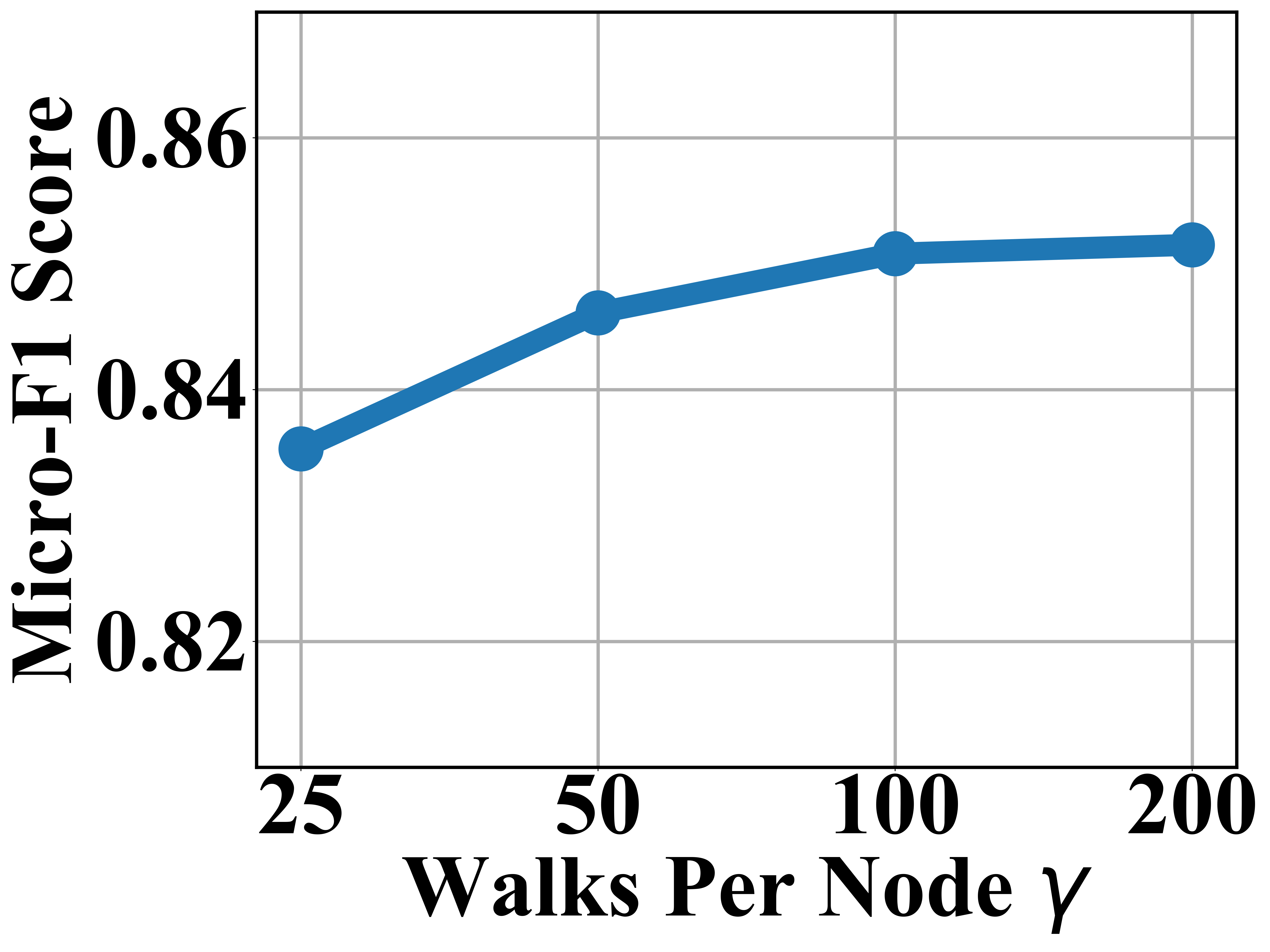}}
%\hspace{0.5cm}
\subcaptionbox{Anonym. walk length\label{fig:param_l}}{\includegraphics[width=0.18\textwidth]{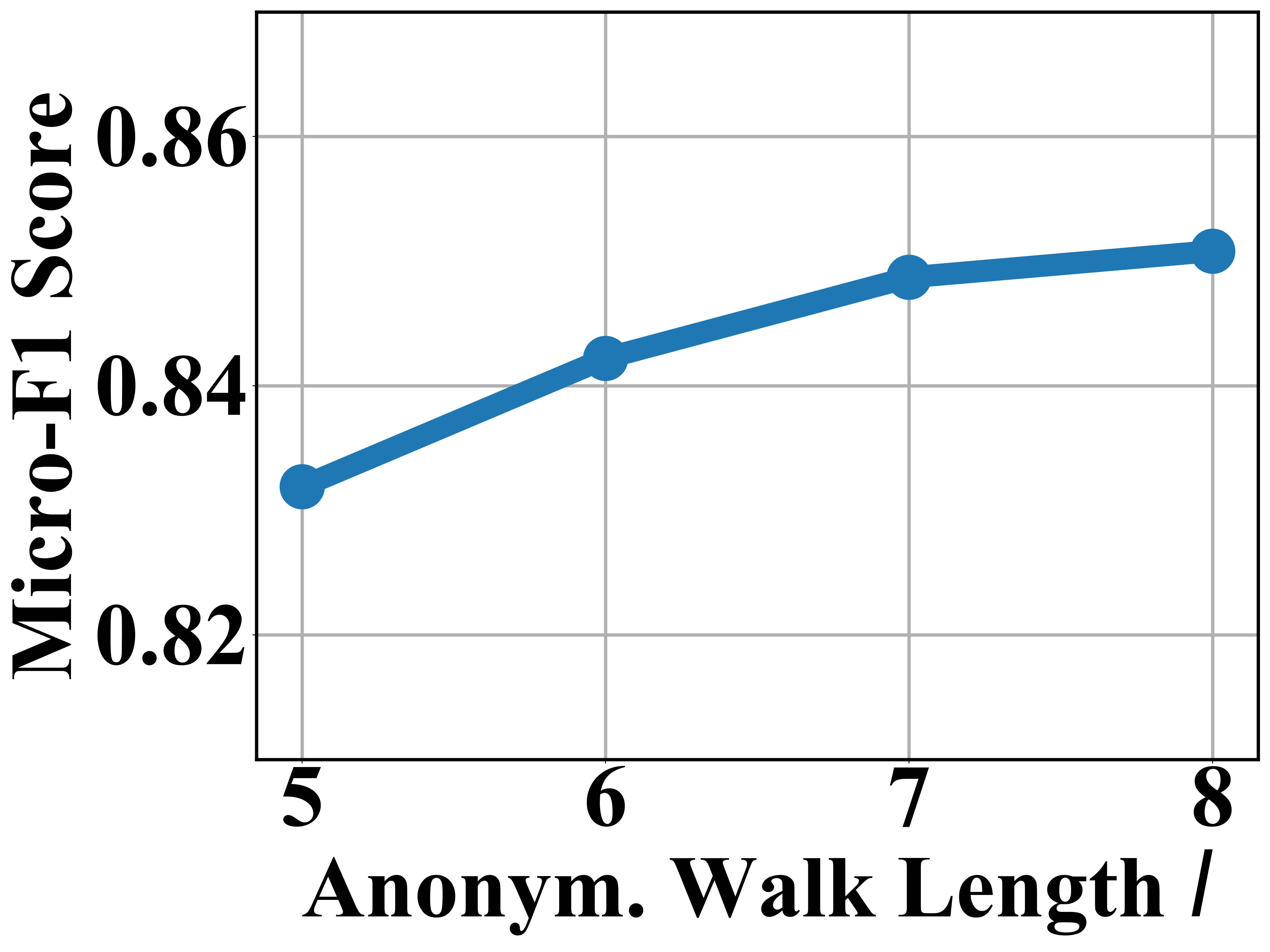}}
%\hspace{0.5cm}
\subcaptionbox{Walk loss weight\label{fig:param_mu}}{\includegraphics[width=0.18\textwidth]{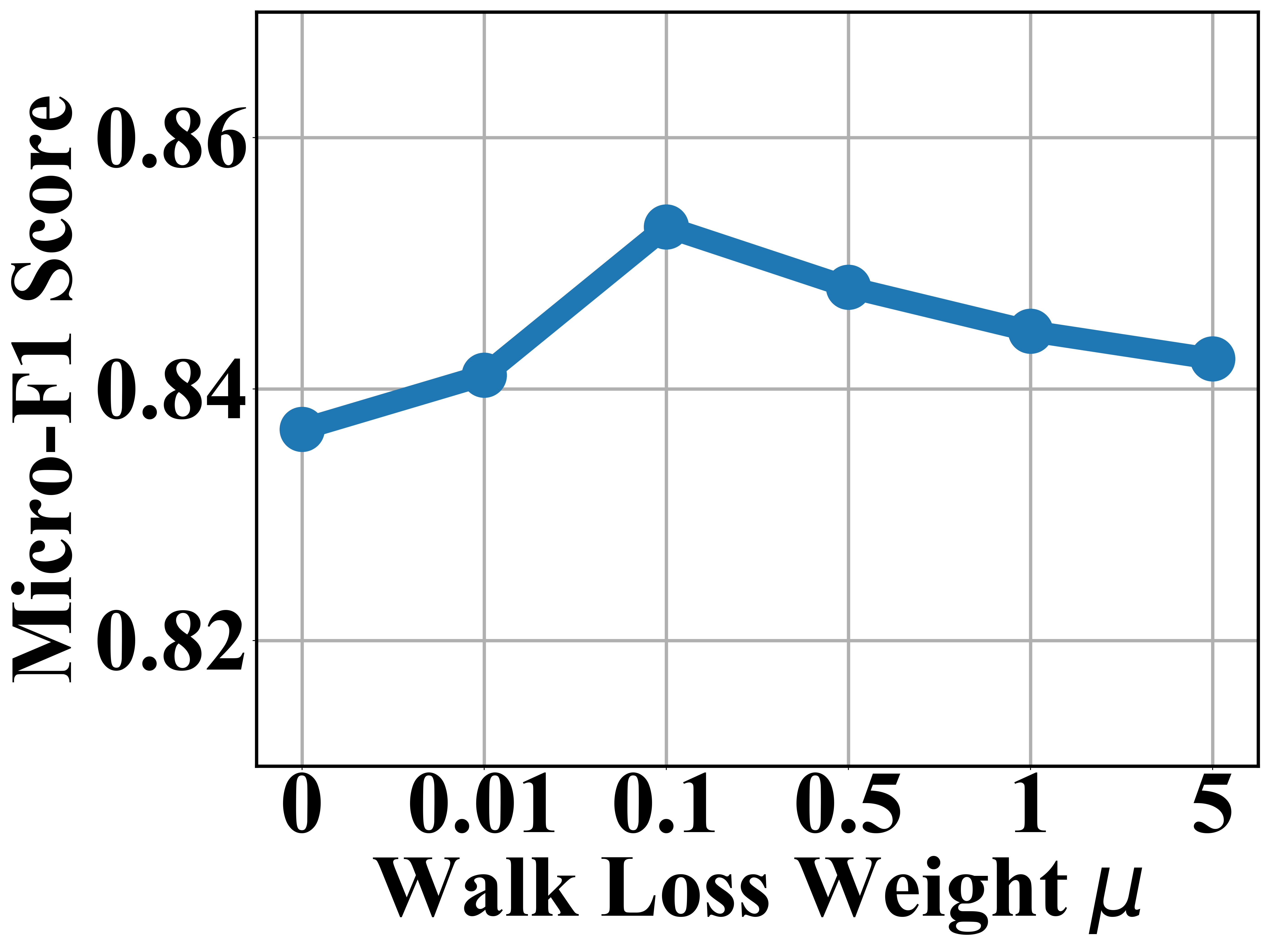}}
%\hspace{0.5cm}
\subcaptionbox{Aggregation scheme\label{fig:comp_aggregation}}{\includegraphics[width=0.18\textwidth]{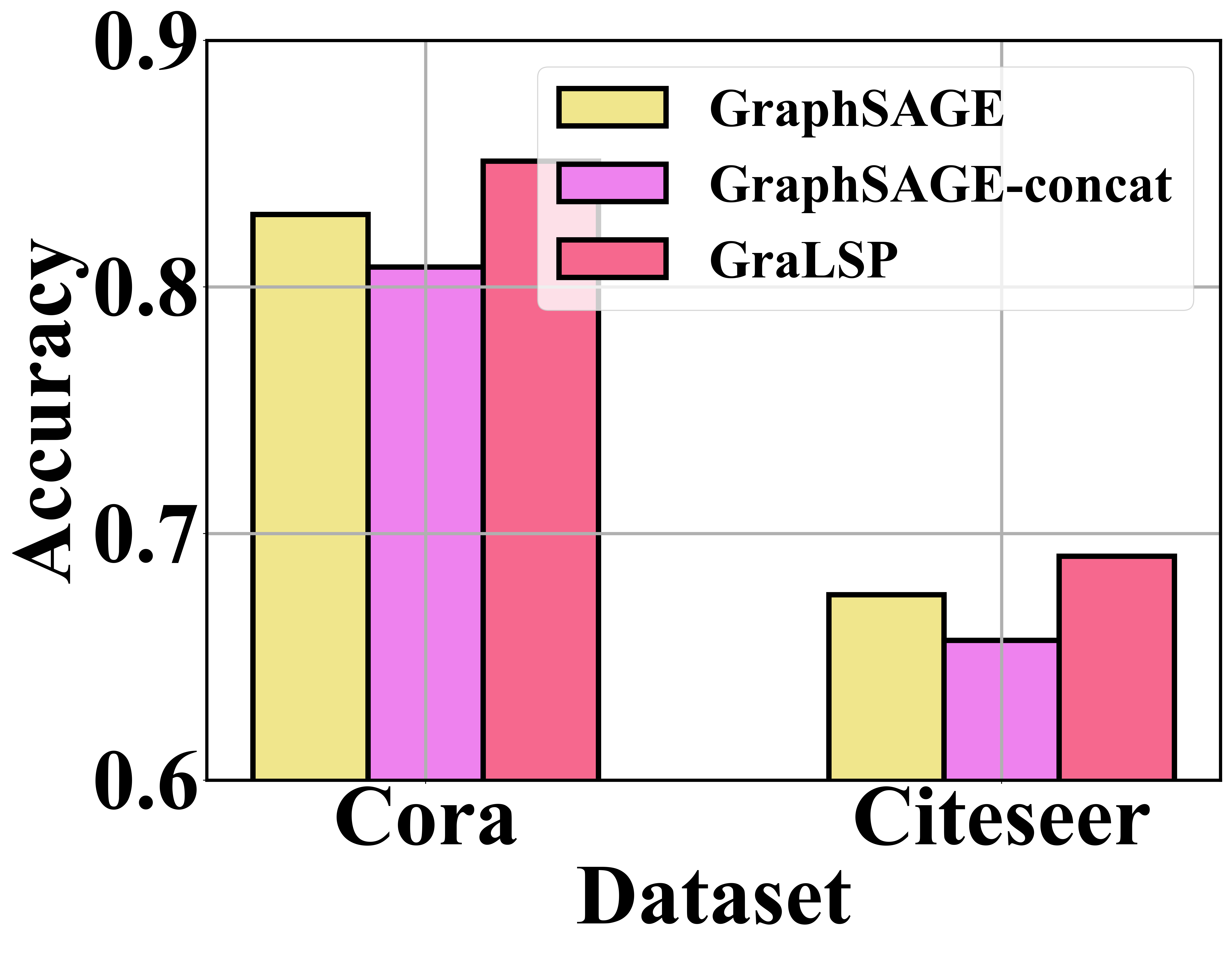}}
\subcaptionbox{Scalability\label{fig:scalability}}{\includegraphics[width=0.18\textwidth]{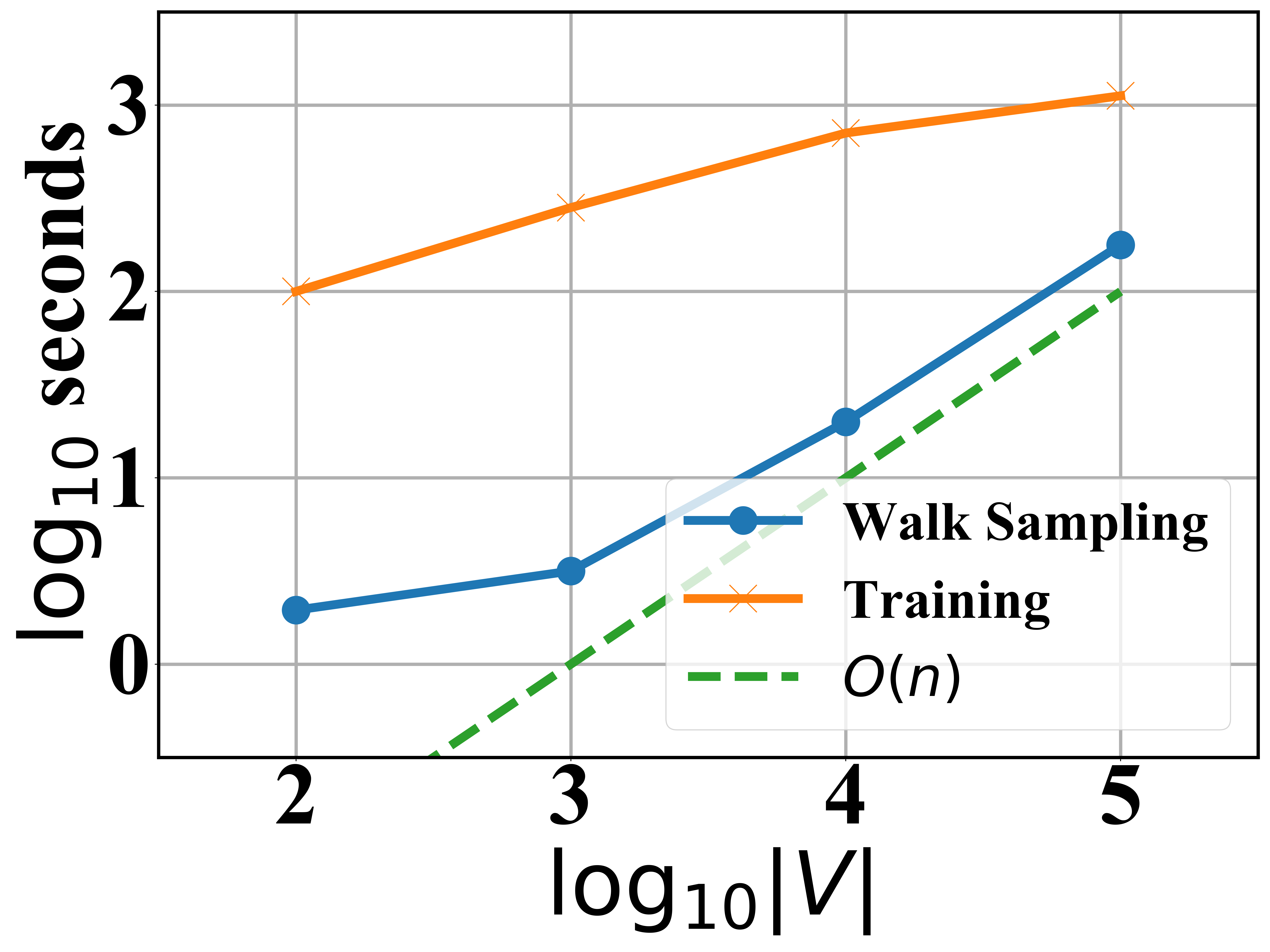}}
\caption{Results of Model Analysis} 
\label{fig:param}
\end{figure*}

\begin{figure*}[t]
\centering
\subcaptionbox{DeepWalk\label{fig:vis_deepwalk}}{\includegraphics[width=0.2\textwidth]{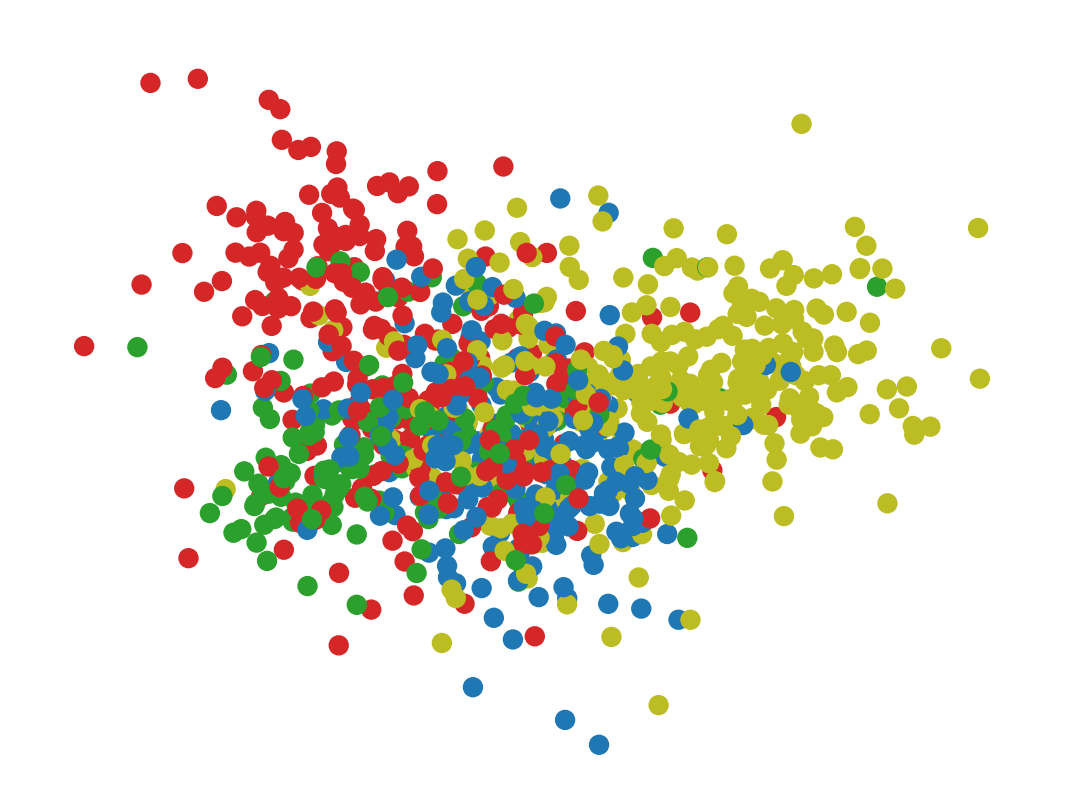}}
\hspace{0.5cm}
\subcaptionbox{struc2vec\label{fig:vis_struc2vec}}{\includegraphics[width=0.2\textwidth]{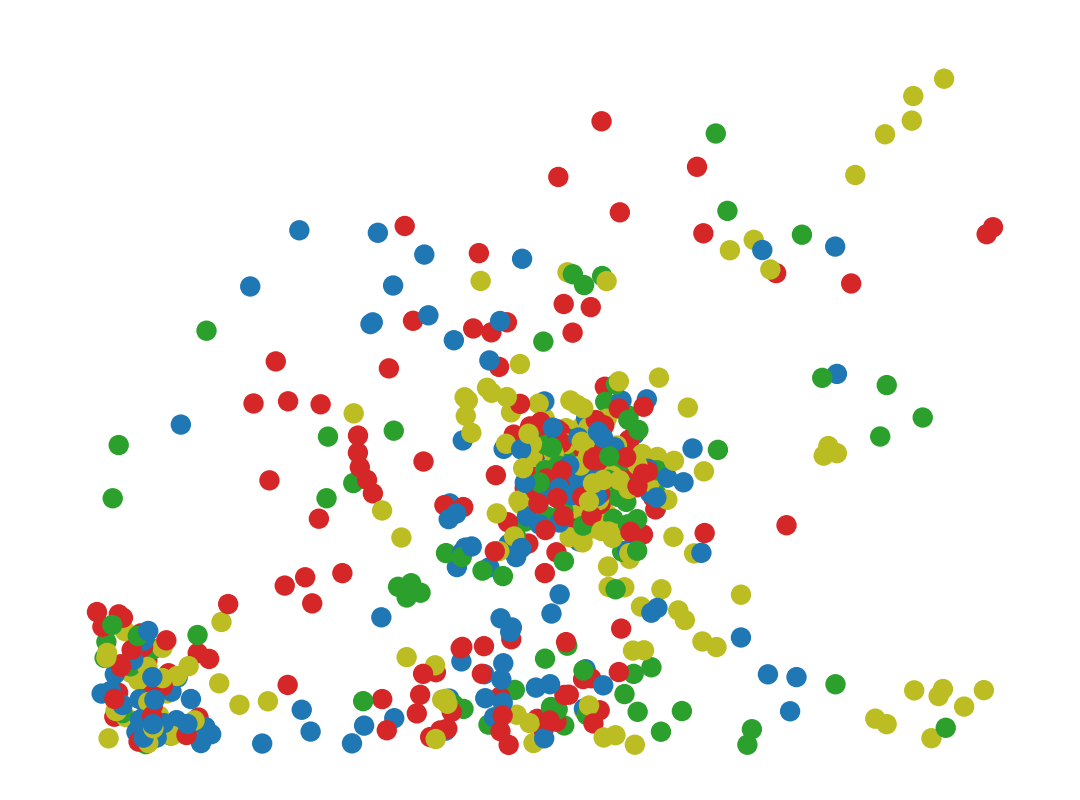}}
\hspace{0.5cm}
\subcaptionbox{GraphSAGE\label{fig:vis_graphsage}}{\includegraphics[width=0.2\textwidth]{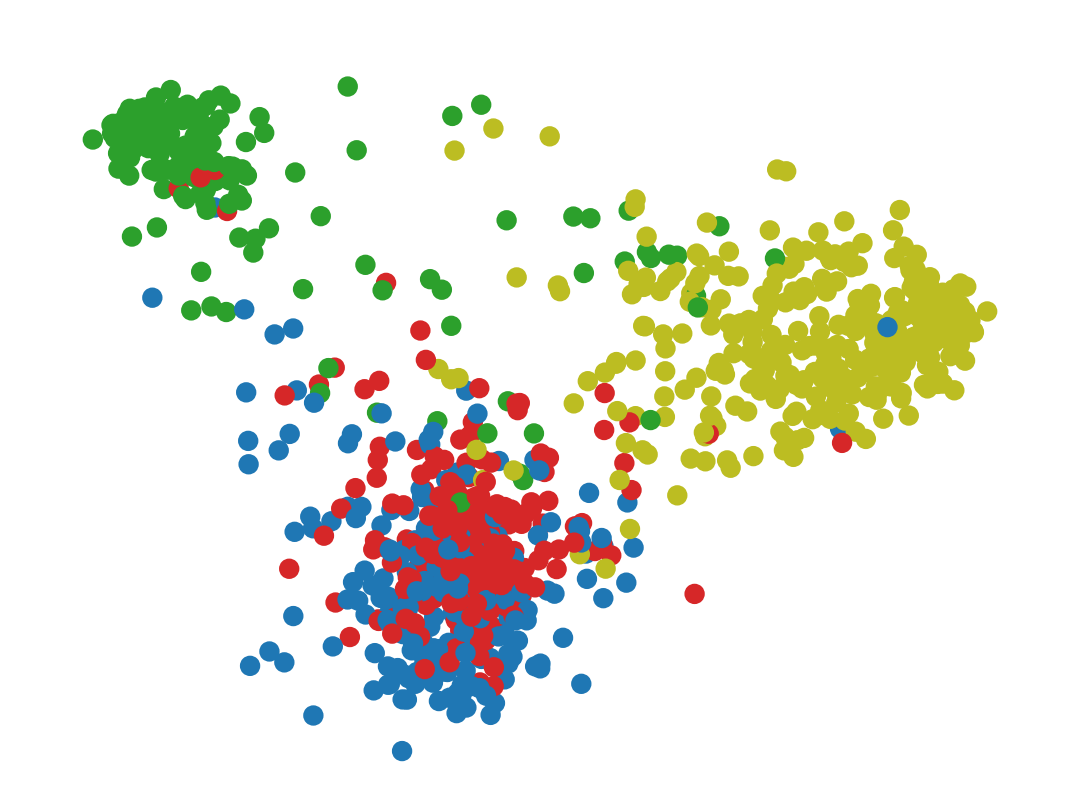}}
\hspace{0.5cm}
\subcaptionbox{GraLSP\label{fig:vis_gralsp}}{\includegraphics[width=0.2\textwidth]{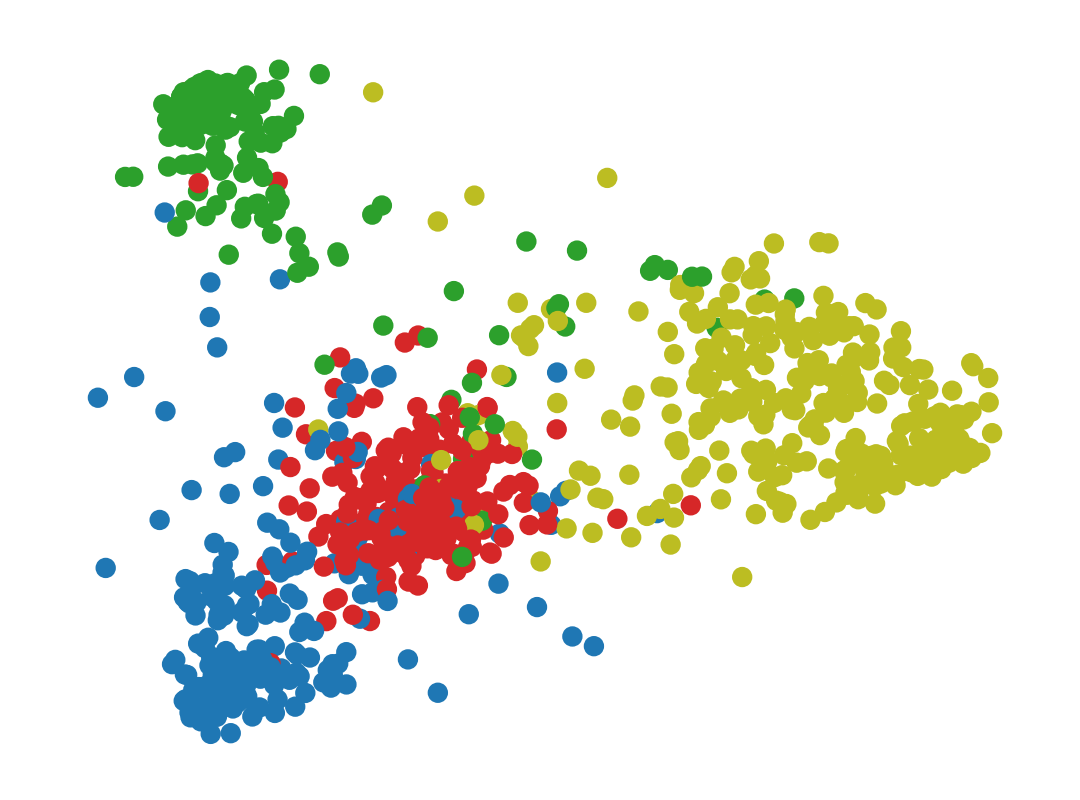}}
\caption{Visualization of representation vectors from various algorithms in 2D space} 
\label{fig:vis}
\end{figure*}

\subsection{Link Prediction}
We then carry out link prediction under the same settings. %It should be noted that because struc2vec and GraphWave do not adopt the similarity defined by links, they are not tested for link prediction. 
We generate the test set by sampling 10\% of the edges as positive edges, which are removed during training, with an identical number of random negative edges. For an edge $\langle i, j\rangle$, we take the inner product of their vectors $\mathbf{h}_i^T\mathbf{h}_j$, which will serve as the score for ranking. We take AUC and recall at 50\% (equal to the number of positive edges) as metrics.

The results are shown in Table \ref{tab:link_prediction}. It can be shown that our model is able to achieve gains compared to GCN, GraphSAGE and GAT, which should not be surprising given that structural patterns will shed light on possible edges \cite{huang2015triadic}, which are better captured by our model. Again, it is not surprising that struc2vec and Graphwave fail to generate satisfactory performances in that they assign similar representations to structurally similar nodes instead of connected nodes.  As for US-Airport dataset, it is likely that local proximity is sufficient to reconstruct the graph, as shown by all baselines except LINE fail to perform well.

\subsection{Model Analysis}
We carry out tests on our model itself, including parameter analysis and scalability. Unless specified, we use node classification on Cora to reflect the performance of the model. All parameters are fixed as mentioned except those tested. 

\subsubsection{Number of Walks Sampled} 
We run GraLSP with number of walks per node $\gamma = 25, 50, 100, 200$, and report their performances in Fig. \ref{fig:param_gamma}. It can be shown that the more walks sampled each node, the better the performance. Empirically, as increasing $\gamma$ from 100 to 200 yields no significant gain, we conclude that $\gamma = 100$ is reasonable in practice. 

\subsubsection{Length of Anonymous Walks} As longer walks are considered, more complex structural patterns are incorporated. 
We take $l = 5, 6, 7, 8$ and show the performances in Fig. \ref{fig:param_l}. As shown, performance improves along with $l$, with decreasing marginal gains. As the number of anonymous walks grows exponentially with $l$, we conclude that $l = 8$ would be sufficient in balancing efficiency and performance. 

\subsubsection{Weight of Objective Functions} We analyze the weight of losses $\mu$, which determines the trade-off between the multi-task objective. We take $\mu=0, 0.01, 0.1, 0.5, 1, 5$, and plot the performances in Fig. \ref{fig:param_mu}. It can be observed that starting from $\mu = 0$, using only the objective in Eqn. \ref{eqn:node}, the performance peaks at $\mu = 0.1$, before taking a plunge afterwards. We hence conclude that, the incorporation of our multi-task objective does enhance the performance of the model.
\subsubsection{Study of Aggregation Scheme} We analyze our aggregation scheme to verify that it enhances the aggregation of features. We compare our model with ordinary GraphSAGE, along with another GraphSAGE with node features concatenated with the distribution of anonymous walks, which proves to be a valid measure of structures (Ivanov \shortcite{ivanov2018anonymous}). We quote this variant as \textit{GraphSAGE-concat}.

We show the results on Cora and Citeseer in Fig. \ref{fig:comp_aggregation}. As shown, with node features concatenated with structural features, the GraphSAGE-concat did not even outperform GraphSAGE, which demonstrates that simply combining them would compromise both. By comparison, our model with adaptive receptive radius, attention and amplification outperforms both GraphSAGE and GraphSAGE-concat. 

\subsubsection{Scalability} We finally analyze the scalability of our model. We run our model on Erdos-Renyi random graphs $G_{np}$ with $n = 100, 1000, 10000, 100000$ and $np=6$. We tested the time needed for the preprocessing (i.e. sampling random walks) and training to converge, which is defined by the loss not descending for 10 continuous iterations.

We plot the time needed with respect to the number of nodes in \textbf{log-log} scale in Fig. \ref{fig:scalability}. As can be seen, both the preprocessing and the training time are bounded by an $O(n)$ complexity, which endorses the scalability of our model. 

\subsection{Visualization on Real World Datasets}
We finally carry out visualization on real world datasets to qualitatively evaluate our model. We learn the representation vectors on Cora, which are then reduced to 2-dimensional vectors using PCA. We select three representative models: DeepWalk (skip-gram), GraphSAGE (GNNs) and struc2vec (structure models), along with our model to compare. 

The plots are shown in Fig. \ref{fig:vis}, where yellow, green, blue and red dots correspond to 4 labels within Cora. As shown, struc2vec (Fig. \ref{fig:vis_struc2vec}) illustrates no clusters as connected nodes do not share similar representations. In addition, while DeepWalk (Fig. \ref{fig:vis_deepwalk}), GraphSAGE (Fig. \ref{fig:vis_graphsage}) and GraLSP (Fig. \ref{fig:vis_gralsp}) all illustrate clustering among nodes with the same label, GraLSP generates clearer boundaries than DeepWalk and GraphSAGE (between blue and red dots).

\section{Conclusion}
We present a GNN framework incorporating local structural patterns to current GNNs, called \textbf{GraLSP}. We start by analyzing drawbacks of current GNNs in identifying local structural patterns, like triads. We then show that anonymous walks are effective alternatives in measuring local structural patterns, and represent them with vectors, which are incorporated into neighborhood aggregation with multiple modules. In addition, we present a multi-task objective preserving proximity between both pairwise nodes and walks. By adequately taking local structural patterns into account, our method outperforms several competitive baselines.

For future work, we plan to extend this paper to GNNs with more sophisticated architectures and more elaborate representations of local structures. In addition, interpretations of structures in GNNs will definitely improve our insight on various network phenomena.

\section*{Acknowledgement}
We are grateful to Lun Du and Yizhou Zhang for their helpful suggestions. This work was supported by the National Natural Science Foundation of China (Grant No. 61876006 and No. 61572041).

\fontsize{9pt}{10pt} \selectfont
\bibliographystyle{aaai}
\bibliography{7432}
\end{document}